\documentclass[letterpaper]{article} 
\usepackage{aaai2026}  
\usepackage{times}  
\usepackage{helvet}  
\usepackage{courier}  
\usepackage[hyphens]{url}  
\usepackage{graphicx} 
\usepackage{natbib}  
\usepackage{caption} 
\usepackage{algorithm}
\usepackage{algorithmic}
\usepackage{pifont}
\usepackage{booktabs}
\usepackage{amssymb}
\usepackage{lineno} 
\usepackage[dvipsnames]{xcolor}
\usepackage{newfloat}
\usepackage{listings}
\DeclareCaptionStyle{ruled}{labelfont=normalfont,labelsep=colon,strut=off} 
\floatstyle{ruled}
\newfloat{listing}{tb}{lst}{}
\floatname{listing}{Listing}
\nocopyright

\title{System~2 Reasoning for Human-AI Alignment: \\ Generality and Adaptivity via ARC-AGI}
\author{
    Sejin Kim,
    Sundong Kim
}
\affiliations{
    Department of AI Convergence \\
    Gwangju Institute of Science and Technology \\
    Gwangju, Republic of Korea \\
    \{sejinkim, sundong\}@gist.ac.kr
}

\usepackage{bibentry}

\begin{document}

\maketitle

\begin{abstract}
Despite their broad applicability, transformer-based models still fall short in System~2 reasoning, lacking the generality and adaptivity needed for human-AI alignment.
We examine weaknesses on ARC-AGI tasks, revealing gaps in compositional generalization and novel-rule adaptation, and argue that closing these gaps requires overhauling the reasoning pipeline and its evaluation.
We propose three research axes:
(1) Symbolic representation pipeline for compositional generality,
(2) Interactive feedback-driven reasoning loop for adaptivity, and
(3) Test-time task augmentation balancing both qualities.
Finally, we demonstrate how ARC-AGI’s evaluation suite can be adapted to track progress in symbolic generality, feedback-driven adaptivity, and task-level robustness, thereby guiding future work on robust human-AI alignment.
\end{abstract}


\section{Introduction}
\label{Sec:Intro}
Artificial Intelligence has recently achieved remarkable capabilities across a wide range of tasks, from language understanding and code generation to image synthesis and scientific discovery~\cite{achiam2023gpt, anthropic2024claude, jumper2021alphafold}. 
Much of this progress has been driven by large language models (LLMs) and foundation models trained on vast internet-scale data. 
These models demonstrate impressive zero-shot and few-shot performance, suggesting a form of generalized competence across modalities.

However, current AI systems remain data-driven. They perform well at interpolation within the training distribution but struggle with systematic generalization to novel, unseen tasks~\cite{sutton2019bitter, chollet2019ARC}.
Their behavior often reflects surface-level patterns rather than deep causal or logical reasoning.
Although LLMs are powerful, they are not AGI.
They lack abstract, compositional, adaptive reasoning to solve unfamiliar problems, respond to evolving goals, or align with human intent in open-ended settings.

One concrete limitation of LLMs is their difficulty with multi-hop reasoning, which requires chaining together multiple pieces of information across steps~\cite{lee2024reasoning, gendron2024large}. 
While LLMs are highly competent at System~1 reasoning, which involves fast, automatic, and pattern-based decision-making, they struggle with System~2 reasoning that demands deliberate, logical, and adaptive thought~\cite{kahneman2011think, anthony2017thinking}. 

We argue that achieving System~2 reasoning is essential for building AI systems that are both generally capable and alignable with human objectives. 
Two interdependent abilities underlie this form of reasoning: \textbf{generality}, which enables models to generalize abstract structure to new settings, and \textbf{adaptivity}, which supports adjustment of reasoning in response to context, feedback, or task novelty~\cite{chollet2019ARC, lake2017building}. 
Without generality, models fail to transfer insights across tasks. Without adaptivity, models fail to respond robustly to distributional shifts or evolving goals. 
These limitations hinder both progress toward AGI and the ability to align AI systems to human values in dynamic environments.

To evaluate these abilities, the Abstraction and Reasoning Corpus (ARC)~\cite{chollet2019ARC} and its successor ARC-AGI benchmark~\cite{rocha2024program} provide structured, high-resolution tests. 
In each ARC task, a model receives 3 to 5 example input–output grid pairs that implicitly define an abstract transformation, and is then asked to infer the correct output for a new test input. 
Success requires generalizing the abstract analogy or transformation and adapting previously learned concepts to apply in a new configuration. 
This makes ARC-AGI a prototypical benchmark for assessing progress on System~2 reasoning.

To date, several approaches have been explored for solving ARC-AGI, including large language models~\cite{butt2024codeit, xu2024llms}, program synthesis~\cite{barke2024hysynth, alford2021neural}, and neuro-symbolic hybrid methods~\cite{lim2024symbolic, hocquette2024relational}. 
However, even the best-performing models currently achieve only around 15\% accuracy on ARC-AGI-2, which underscores both the inherent difficulty of the benchmark and the fundamental limitations of existing reasoning strategies.

We posit that future success on ARC-AGI, and by extension meaningful progress toward AGI, will require models that genuinely implement System~2 reasoning. 
In particular, we identify three key directions: (1) models that prioritize compositional generality, (2) models that enable feedback-driven adaptivity, and (3) hybrid approaches that balance both through test-time flexibility. 
\textbf{This position paper presents a conceptual framing of these three research axes and argues that they provide a viable roadmap for building general, adaptive, and ultimately alignable reasoning systems}.

\begin{figure*}[ht!]
    \centering
    \includegraphics[width=\linewidth]{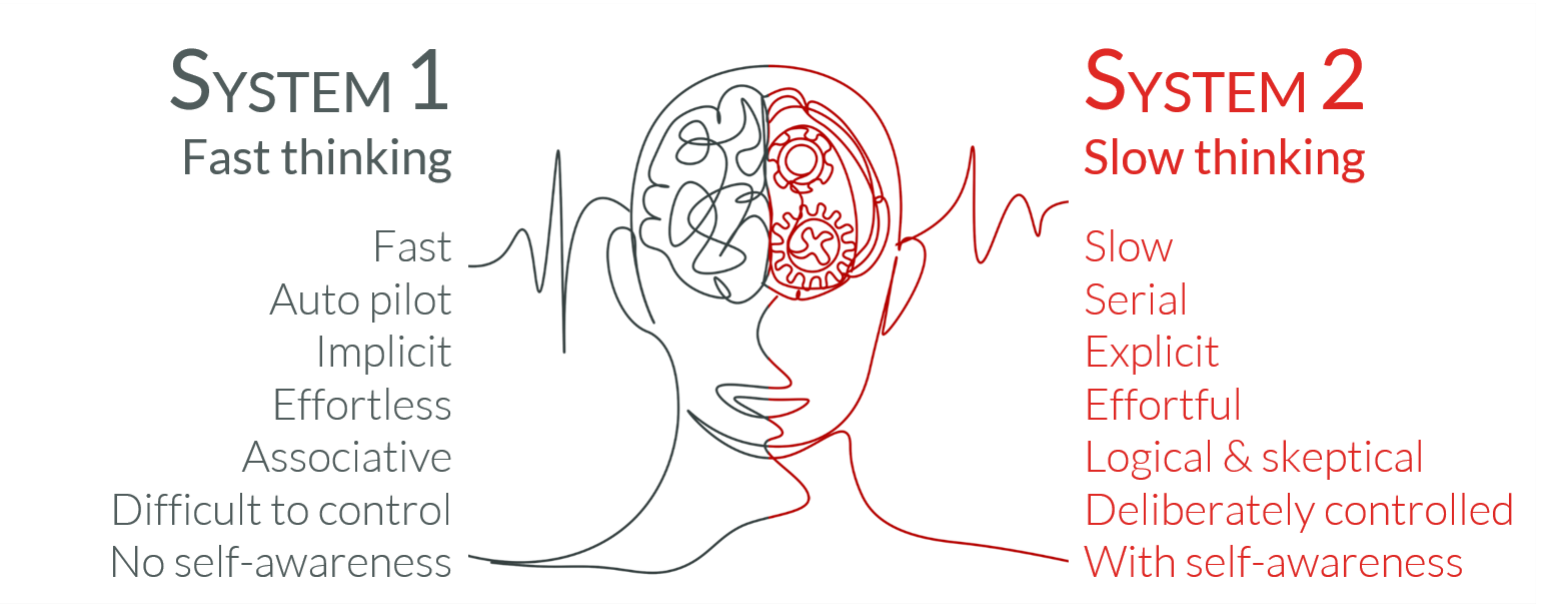}
    \caption{A conceptual comparison between System~1 and System~2 reasoning, highlighting key cognitive characteristics of each mode. System~1 is fast, automatic, and associative, operating without self-awareness or deliberate control. In contrast, System~2 is slow, deliberate, and logical, involving explicit reasoning and self-aware thought. This dichotomy provides the foundational motivation for distinguishing between intuitive (System~1) and reflective (System~2) reasoning.}
    \label{fig:system1-vs-system2}
\end{figure*}

\section{Background}
\label{Sec:Background}

\begin{table*}[ht!]
\centering
\caption{A qualitative comparison of three major approaches commonly used for solving ARC tasks and enabling System~2 reasoning, evaluated in terms of their support for generality and adaptivity. The table summarizes how well each approach satisfies these two core capabilities. Here, \textcolor{Green}{\ding{108}} indicates that the capability is well supported, \textcolor{Goldenrod}{\ding{115}} suggests that it is only partially funded, and \textcolor{Red}{\ding{55}} demonstrates that the ability is not supported at all.}
\begin{tabular}{@{}lccp{6cm}@{}}
\toprule
\textbf{Method} & \textbf{Generalization} & \textbf{Adaptation} & \textbf{Notes} \\
\midrule
Program Synthesis & \textcolor{Goldenrod}{\ding{115}} & \textcolor{Goldenrod}{\ding{115}} & Limited by reliance on pre-defined \textit{DSLs} \\
LLMs & \textcolor{Green}{\ding{108}} & \textcolor{Goldenrod}{\ding{115}} & Improving with \textit{large-scale data} \\
Transformers & \textcolor{Goldenrod}{\ding{115}} & \textcolor{Red}{\ding{55}} & Only effective for \textit{structured} tasks \\
\bottomrule
\end{tabular}
\label{tab:existing_approaches}
\end{table*}

\subsection{Two Modes of Reasoning: System~1 and System~2}
The concepts of System~1 and System~2 reasoning, introduced in cognitive psychology~\cite{kahneman2011think}, describe two distinct modes of thinking. 
System~1 is fast, automatic, and intuitive, relying on pattern recognition and learned experiences. 
Humans often employ System~1 for routine tasks such as driving or basic arithmetic, which require minimal conscious effort. 
In AI, similar behavior can be achieved through neural networks trained for rapid inference, or when combined with search methods to mimic fast decision-making~\cite{anthony2017thinking}. 
However, such models typically excel only in familiar, well-structured environments and struggle to adapt when faced with novel or complex problems.

In contrast, System~2 reasoning is slow, deliberate, and analytical.
It is activated when humans face unfamiliar challenges that demand logical deduction, abstraction, and planning.
This mode of reasoning enables the formulation of strategies in dynamic, unpredictable situations and supports capabilities essential for general problem-solving and human-like intelligence.
Incorporating System~2 reasoning into AI would address current limitations in abstraction and logical flexibility~\cite{booch2021thinking}.

Ultimately, the two systems differ not only in speed and effort but also in their underlying mechanisms.
System~1 relies on implicit, associative processes, while System~2 engages explicit, rule-based reasoning.
For AGI, the integration of both modes is essential: System~1 supports efficient processing in familiar domains, while System~2 enables generalization and adaptation in novel settings.
Together, they form a complementary framework for building AI systems capable of robust, human-aligned reasoning.
\subsection{Generality and Adaptivity: Two Key Components of System~2 Reasoning}

System~2 reasoning enables deep, logical thought processes for handling novel and complex tasks. 
To fully realize its potential in AI systems, two critical capabilities are essential: \textit{generality} and \textit{adaptivity}. 
Together, they define the reasoning flexibility required for Artificial General Intelligence (AGI). 
An AGI system must not only perform well on familiar tasks but also generalize to unseen problems and adapt to new environments without extensive retraining.

\textbf{Generality} allows AI models to apply prior knowledge to novel contexts. 
Rather than overfitting to training data, a general AI system can abstract underlying principles and transfer them across tasks or domains. 
For example, a model trained on object recognition should be able to extend its understanding to new categories with minimal supervision.
The generality of an AI model $\mathcal{M}$ over a task set $\mathcal{T}$ and domain knowledge $\mathcal{K}$ can be formalized as:

\begin{equation}
\label{Eq:generality}
G(\mathcal{M}, \mathcal{T}, \mathcal{K}) = \frac{1}{N} \sum_{i=1}^{N} P(T_i|\mathcal{K})
\end{equation}

Here, $P(T_i|\mathcal{K})$ denotes the probability that the model successfully solves a novel task $T_i$ using its acquired knowledge $\mathcal{K}$. 
This metric reflects how well the model generalizes across diverse tasks without task-specific tuning.

\textbf{Adaptivity}, in contrast, refers to the model’s ability to modify its behavior in response to changing environmental conditions. 
It is essential for maintaining robustness and relevance in dynamic or unpredictable scenarios.
For instance, a robot that operates in multiple physical environments must adjust its policy according to different layouts or obstacles. 
Formally, we define adaptivity as:

\begin{equation}
\label{Eq:adaptivity}
A(\mathcal{M}, \mathcal{T}, \mathcal{K}, \mathcal{E}) = \frac{1}{N} \sum_{i=1}^{N} P(T_i|\mathcal{K}, E_i)
\end{equation}

In this formulation, $\mathcal{E} = \{E_1, E_2, \cdots, E_N\}$ represents varying environmental conditions. 
$P(T_i|\mathcal{K}, E_i)$ is the probability of solving task $T_i$ given both global knowledge and the specific condition $E_i$.

\textbf{The synergy between generality and adaptivity} is central to AGI-level reasoning.
A brilliant system must abstract from prior experience while also adjusting to contextual shifts, since this dual capability enables robust reasoning in unfamiliar or evolving tasks.
Therefore, enhancing generality and adaptivity is a prerequisite for bridging the gap between current narrow AI systems and AGI.
\subsection{ARC: A Reasoning-Centric Benchmark for AGI Research}
The Abstraction and Reasoning Corpus (ARC)~\cite{chollet2019ARC} was introduced as a diagnostic benchmark for evaluating AI systems' ability to reason in a human-like way.
Each ARC task presents a few pairs of input and output grids, and the model must predict the correct output for a new input grid using only these limited demonstrations.
No prior training on the tasks is allowed, and the use of large datasets or handcrafted features is explicitly discouraged.
As a result, solving ARC tasks demands the ability to infer abstract rules and apply them flexibly to unseen examples.

What makes ARC especially challenging is that it is designed to assess cognitive functions that go beyond statistical pattern recognition.
The tasks often require compositional reasoning, symbolic manipulation, and analogy-making, which are capabilities central to what psychologists refer to as System~2 reasoning.
Models must identify underlying transformations, manipulate high-level concepts, and generalize from sparse data in ways that mirror human reasoning under uncertainty.

ARC also serves as a practical testbed for measuring the two key dimensions of AGI reasoning emphasized in this paper: \textbf{generality} and \textbf{adaptivity}.
Generality is needed because the tasks are diverse and unpredictable; a model must reuse its knowledge flexibly across different settings.
Adaptivity is equally essential, as each ARC task introduces new structural rules, and the model must quickly adjust its strategy based on limited examples.
Thus, strong performance on ARC indicates not just surface-level pattern matching but the presence of deeper abstraction and flexible reasoning.

In this paper, we propose that ARC should be viewed not only as a benchmark, but also as a principled microcosm of AGI-level reasoning. 
Its design integrates the challenges of generality, adaptivity, and symbolic abstraction, all of which are central to System~2 cognition. 
By explicitly targeting these capacities, ARC enables the evaluation of AI systems that aim to achieve more robust, general-purpose reasoning. 
Progress on ARC is therefore not only a matter of empirical performance, but a meaningful step toward developing AI systems capable of adaptive, human-aligned intelligence.

\section{Challenges of System~2 Reasoning in ARC}

Recent efforts to solve ARC tasks have employed various paradigms, including program synthesis, large language models (LLMs), and specialized transformer architectures.  
From a System~2 reasoning perspective, these approaches have contributed to improved generalization but continue to struggle with adaptivity, especially when faced with unfamiliar rules or sparse demonstrations.  
Table~\ref{tab:existing_approaches} provides a comparative summary of these methods, evaluating their strengths and weaknesses across the dimensions of generality, adaptivity, and System~2 capabilities.  
This analysis highlights the need for new approaches that better integrate both generalization and adaptation to advance reasoning in ARC.  
In the following subsections, we examine the specific limitations of program synthesis, LLM-based methods, and transformer variants in greater detail.

\subsection{Program Synthesis: Constrained by Pre-defined DSLs in Generality and Adaptivity}

Program synthesis approaches aim to generate interpretable programs that solve tasks based on input-output examples.  
In the context of ARC, this paradigm is particularly attractive due to its support for symbolic reasoning and transparency.  
However, ARC tasks demand not just symbolic reasoning but also a high degree of generality and adaptivity, qualities that current synthesis methods struggle to achieve.  
Despite advances in neural guidance, object-centric reasoning, and logical decomposition, program synthesis remains fundamentally limited by its dependence on a predefined domain-specific language (DSL).  
This reliance inherently restricts the expressivity and flexibility of the system.

Neural program synthesis uses deep learning to guide search within a DSL~\cite{acquaviva2022communicating, ainooson2023approach, alford2021neural, banburski2020dreaming, barke2024hysynth, ferre2024tackling, hocquette2024relational, lei2024generalized, lim2024symbolic, rocha2024program, xu2023graphs}.  
These methods improve efficiency or interpretability, such as using bidirectional search~\cite{alford2021neural} or explicitly communicating concepts~\cite{acquaviva2022communicating}.  
Object-centric synthesis improves generality by reasoning over object abstractions and relationships~\cite{ferre2024tackling, lei2024generalized}, while symbolic or inductive logic programs increase rule-based generalization across examples~\cite{barke2024hysynth, rocha2024program}.  

Nevertheless, all these methods share a critical limitation in that they operate within the boundaries of an initial DSL.  
If a task requires operations or abstractions not expressible within the DSL, the system fails to generalize or adapt.  
This brittleness is especially problematic for ARC, where tasks are intentionally diverse and the required transformations are not known in advance.

As reflected in Eq.~\ref{Eq:generality} and Eq.~\ref{Eq:adaptivity}, the DSL effectively defines the domain knowledge $\mathcal{K}$ available to the system.  
The fixed scope of $\mathcal{K}$ constrains both $P(T_i|\mathcal{K})$ and $P(T_i|\mathcal{K}, E_i)$, which represent the model's ability to generalize across tasks and adapt to varying environments.  
To improve ARC performance, future program synthesis research must focus on developing more expressive and adaptive DSLs or mechanisms that allow the DSL to evolve dynamically in response to task requirements.
\subsection{LLMs Demonstrate Strong Generality but Face Adaptivity Constraints}
\label{sec:LLMs}

Large Language Models (LLMs) have been successfully applied to a variety of tasks, including abstract reasoning problems in the Abstraction and Reasoning Corpus (ARC).  
These models excel at recognizing patterns and generating solutions by leveraging large-scale pretraining.  
In ARC, LLMs have been used to perform inductive reasoning, symbolic conversion, and hypothesis refinement.  
Although they demonstrate strong generality by adapting knowledge across tasks, they still face challenges when encountering novel problems that require deeper and flexible adaptation.

Recent studies have explored diverse strategies for applying LLMs to ARC~\cite{butt2024codeit, galanti2024intelligence, hu2024automated, mirchandani2023large, qiu2024phenomenal, shin2024mclarc, tan2023large, wang2024hypothesis, wang2024speak, xu2024llms}.  
For instance, \citet{qiu2024phenomenal} proposed hypothesis refinement, allowing LLMs to iteratively improve their outputs based on feedback.  
\citet{shin2024mclarc} translated symbolic tasks into natural language explanations to facilitate reasoning, although their method struggled with highly abstract rules.  
\citet{mirchandani2023large} demonstrated LLM-based pattern recognition, yet the models remained constrained by their reliance on pretraining and lacked the flexibility to adapt strategies dynamically.

While LLMs show strong generality in ARC tasks, this largely stems from their broad training data ($\mathcal{K}$).  
Consequently, they often interpolate within seen patterns rather than engage in genuine reasoning, and falter when faced with novel, untrained structures~\cite{lee2024reasoning}.  
This raises a key concern: current LLMs may not reason abstractly but instead rely on memorized statistical patterns.

Recent improvements in LLM-based ARC performance often rely on increasing model scale and pretraining data, rather than fundamentally more adaptive reasoning mechanisms~\cite{openai2024o1}.  
This reflects a general trend of enhancing generality through data expansion while leaving the problem of on-the-fly adaptation largely unresolved.

To address this, OpenAI's o1 model integrates Chain-of-Thought (CoT) reasoning to support multi-step problem-solving~\cite{openai2024o1}.  
This design aims to improve adaptability by encouraging iterative thinking and flexible strategy shifts~\cite{marino2024fast}.  
Although o1 demonstrates potential, its performance on ARC tasks and its contribution to AGI-level reasoning remain inconclusive~\cite{o12024arc}.  
Future work must assess whether such models can truly bridge the gap between data-driven generalization and adaptive, System~2 reasoning.

\subsection{Structure-Biased Transformers for ARC}

Transformers have demonstrated remarkable capabilities in fields such as language modeling and vision tasks. 
In the context of ARC, several studies have explored the use of transformer variants that incorporate explicit inductive biases to enhance performance on structured reasoning tasks. 
These models show moderate generalization, particularly in domains where structural regularities such as symmetry, object relations, or counting play a dominant role. 
However, they often struggle to adapt to novel, unstructured tasks, limiting their flexibility and broader applicability.

Various transformer-based models have been developed with structural priors to address specific sub-problems in ARC.  
For instance, Atzeni et al.~\cite{atzeni2023infusing} introduced symmetry-aware attention mechanisms to help models generalize across geometrically consistent patterns.  
Park et al.~\cite{park2023unraveling} proposed object-centric transformers that improve generalization by focusing on object relationships and interactions.  
Ouellette et al.~\cite{ouellette2023counting} enhanced transformer architectures with specialized counting modules for quantitative reasoning.  
These designs are effective in tasks aligned with their structural assumptions, but they tend to falter when applied to ARC problems with novel or abstract rules outside their predefined priors.

From a System~2 reasoning perspective, these structure-biased transformers are limited in their adaptivity.  
Their performance is often constrained by the specific inductive biases embedded during model design, which makes them inflexible in scenarios where new reasoning strategies must be inferred on the fly.  
Although their inductive priors can enhance performance in well-structured tasks, this very specialization hampers their ability to generalize across the full diversity of ARC tasks.

As summarized in Table~\ref{tab:existing_approaches}, transformer variants offer a partial solution to the ARC challenge by improving generalization in structured domains.  
However, their lack of adaptivity underscores the need for more flexible and compositional reasoning architectures that can operate beyond the assumptions hard-coded into the model.  
Bridging this gap is essential for progress toward AGI-level reasoning.

\section{Toward System~2 Reasoning in ARC: Promising Directions}

The preceding analysis reveals that while existing ARC-solving approaches offer partial generalization, they consistently fall short in adaptive reasoning, which is a critical trait of System~2 cognition. 
Most current models can leverage structural priors or large-scale data to generalize within familiar distributions, but they struggle to respond flexibly to novel or dynamically shifting tasks. 
This limitation underscores the need for fundamentally new approaches that are not only capable of identifying patterns but also of reasoning through uncertainty and adapting to unforeseen challenges.

To address this, we propose three interrelated research directions that align with the demands of AGI-level reasoning and the structure of the ARC benchmark. 
Each direction targets a different axis of the generality and adaptivity space, aiming to reinforce the core pillars of System~2 thinking: abstraction, flexibility, and strategic reasoning.

First, we advocate for a \textit{symbolic representation pipeline}, which promotes compositional generality by transforming low-level perceptual data into structured, high-level symbolic abstractions. 
Such a pipeline facilitates reasoning over abstract representations that are transferable across tasks, enabling models to decompose problems semantically and recombine known components in novel ways.

Second, we introduce an \textit{interactive, feedback-driven reasoning loop} designed to support real-time adaptivity. 
Rather than treating inference as a one-shot process, this loop allows models to iteratively refine their hypotheses through feedback signals, whether from the environment, internal self-evaluation, or external correction. 
This approach mirrors the human capacity for trial, error, and strategic adjustment.

Third, we highlight the importance of \textit{test-time task augmentation and training}, a strategy that dynamically alters task representations or generates task variants during inference. 
By exposing models to perturbed or enriched task scenarios at test time, this method improves robustness to distribution shifts and enables models to generalize more effectively to unfamiliar patterns and constraints.

Together, these three directions form a synergistic framework for System~2 reasoning: abstraction through symbolic representation, flexibility through interactive reasoning, and robustness through dynamic task adaptation. 
By pursuing these avenues in tandem, we believe AI systems can be made more capable of deep abstraction, strategic flexibility, and resilient problem-solving, which are the hallmarks of true System~2 reasoning.

\subsection{Symbolic Task Abstraction for Generalizable Reasoning}

To achieve System~2 reasoning in ARC and similar domains, a model must not only process input data but also extract and operate over abstract, symbolic representations that capture high-level task semantics. One promising direction is symbolic task abstraction, which transforms visual or structured data into interpretable representations that support generalizable reasoning across diverse tasks. By explicitly modeling task structures, such abstractions reduce reliance on pixel-level patterns and shift AI reasoning from surface-level correlations to deeper conceptual understanding.

Several recent works demonstrate tangible progress in this direction. ARC-KG~\cite{lim2024symbolic} constructs a scene-graph-style knowledge graph from ARC input-output grids, where each node represents an object and each edge encodes relational or spatial transformations. This symbolic form enables an abductive solver to reason over the graph, iteratively inferring transformation rules that best explain the observed behavior. By encoding tasks into a shared symbolic structure, ARC-KG provides a reusable scaffold that supports generalization across structurally similar tasks. Notably, this approach reflects the human strategy of decomposing problems into semantic components, allowing for cross-task analogy and abstraction.

Similarly, PeARL~\cite{bober2024neural} proposes a domain-specific language (DSL) to semantically abstract ARC tasks into interpretable programs. The DSL includes constructs for handling visual concepts such as shape, color, motion, symmetry, and alignment. This allows a neural network to learn semantic primitives that are reusable across tasks, effectively compressing knowledge into a symbolic form. PeARL emphasizes compositionality, enabling models to assemble known concepts into new configurations and thereby generalize more effectively to unseen tasks. This mirrors cognitive flexibility in humans, where known subskills can be recombined to solve new problems.

Other approaches explore abstraction through latent or compressed structures. CompressARC~\cite{liao2025compress} introduces a multi-tensor architecture that encodes various levels of task information in a compact, hierarchical format. These latent tensors are designed to capture not only local patterns but also abstract properties such as symmetry or repetition, enabling efficient downstream reasoning. Likewise, Latent Program Networks (LPN)~\cite{macfarlane2025searching} embed entire ARC tasks into a latent program space, where reasoning proceeds via learned search over the task embeddings. This allows the model to operate in an abstract representation space rather than directly over pixel data, making reasoning more scalable and generalizable.

Another promising direction within symbolic abstraction is to learn high-level symbolic representations directly from human problem-solving behavior.
Rather than engineering symbolic structures manually, this approach seeks to infer latent cognitive constructs—such as goals or intentions—by analyzing human trajectories.
For instance, IntentionLearning~\cite{kim2025addressing} frames ARC problems as sequences of high-level intentions extracted from human demonstrations, instead of low-level action steps.
These intentions often reflect abstract concepts such as object grouping, symmetry enforcement, or color alignment, which are not only symbolic in nature but also transferable across tasks.
By learning from these representations, models can acquire abstract strategies that resemble human reasoning patterns and generalize beyond surface-level input features.
This method complements existing approaches by embedding symbolic structure through experience, potentially offering more scalable and human-aligned abstraction.

These diverse approaches, including handcrafted symbolic graphs, domain-specific languages, latent abstraction mechanisms, and intention-level modeling from human trajectories, share a common objective: to distill task semantics into structured forms that support high-level, cross-task reasoning. Symbolic task abstraction serves as a critical bridge between raw inputs and cognitive inference, aligning with the compositional and deliberative nature of human System2 thinking. Among these, methods such as IntentionLearning\cite{kim2025addressing} show how symbolic structures can also emerge from human reasoning traces, enabling models to internalize abstract strategies that are transferable across tasks. Moving forward, research should pursue more expressive and flexible abstraction frameworks, including hybrid symbolic-neural architectures, hierarchical representations, and graph-based reasoning systems. These directions offer a promising path toward enhancing generalization in abstract reasoning domains and advancing toward AGI-level cognition.

\subsection{Interactive Feedback-Driven Reasoning for Adaptivity}

Achieving true adaptivity in System~2 reasoning requires more than just static inference capabilities. AI systems must be able to iteratively revise their hypotheses or strategies based on feedback—both internal and external—thereby forming a feedback-driven reasoning loop that reflects the human-like process of trial, reflection, and refinement. Such a loop is particularly valuable for tackling ARC tasks, where the solution is not always immediately evident and may require multiple reasoning steps that evolve with feedback.

Recent research has shown promising results in constructing such interactive reasoning loops. For instance, ConceptSearch~\cite{singhal2025conceptsearch} explores program space using LLMs and refines its search based on evaluation feedback. This iterative refinement enables the system to prune implausible candidates and focus on program concepts that are more aligned with the desired output, enhancing adaptivity over time. Similarly, a multi-agent LLM framework~\cite{mirchandani2023large} proposes a collaborative framework where different LLM-based agents adopt diverse reasoning strategies and exchange feedback with one another. This inter-agent dialogue mimics human collaborative problem-solving and supports flexible adaptation when individual agents encounter impasses.

A more targeted approach is seen in CodeIt~\cite{butt2024codeit}, which employs a prioritized hindsight replay mechanism. Here, the model revisits past failed attempts and selectively incorporates feedback from high-quality demonstrations to improve subsequent outputs. This design creates a self-improving loop where the model not only learns from external feedback but also evaluates and learns from its own past experience. Such retrospective learning enhances adaptivity by allowing the model to escape suboptimal reasoning patterns.

Finally, REx~\cite{tang2024code} offers further evidence of feedback-driven adaptivity. It frames code repair as an interactive task, where the model iteratively proposes fixes and evaluates them against feedback signals. Over time, the model balances exploration of novel fixes with exploitation of previously successful patterns, demonstrating a flexible reasoning loop that adapts to error patterns in a principled way.

These studies suggest that interactive feedback loops are a vital component of adaptive reasoning. Rather than relying on a single-shot inference, models that engage in multiple cycles of hypothesis generation, feedback acquisition, and strategy revision are more likely to succeed in novel, uncertain, or structurally complex tasks. This aligns with the human cognitive process of System~2 reasoning, where adaptation is often a result of learning from intermediate outcomes. Future work should investigate how to structure and guide these feedback loops, including which forms of feedback (e.g., symbolic, programmatic, or natural language) are most effective and how they can be integrated into learning and inference pipelines.

\subsection{Test-Time Task Augmentation and Training for Robustness}

A core limitation of many ARC-solving models lies in their inability to adapt dynamically to unfamiliar tasks at inference time. While generalization can be achieved through large-scale training, true adaptivity requires models to update or refine their internal reasoning based on the specific structure of each new task. To address this, recent research has explored the use of test-time training (TTT), a technique in which the model continues to learn or fine-tune during inference through task-specific augmentations or internal feedback.

One pioneering work in this area is MARC~\cite{akyurek2024surprising}, which first introduced TTT to the ARC domain. Instead of solving tasks with a static model, MARC continually refines its predictions through test-time optimization. By perturbing inputs and backpropagating on unsupervised losses such as agreement with augmented samples, MARC demonstrates that even small models can outperform much larger baselines when equipped with adaptive inference capabilities. This result strongly suggests that adaptivity at inference time is just as important as model capacity during training.

Building on this idea, several subsequent models have integrated TTT into more complex pipelines. For example, BARC~\cite{li2025combining} combines inductive and transductive reasoning with test-time self-training to boost performance. The model first generates pseudo-labels using its most confident predictions, and then adapts itself by training on these labels through multiple iterations. This iterative refinement improves its ability to generalize abstract patterns across a task.

Similarly, LLM ARChitect~\cite{franzen2024llm} incorporates TTT into a modular system that combines LLM-based pattern discovery and grid transformations. During inference, LLM ARChitect fine-tunes its modules using test-time augmentations such as slight modifications of input grids or synthetic analogs. This allows the system to refine its initial hypotheses and select the most plausible transformation based on consistency across augmented inputs.

The growing adoption of TTT is also reflected in the ARC Prize 2024 Technical Report~\cite{chollet2024arc}, which highlights it as a common ingredient among top-performing entries. Several top-performing teams in the ARC Prize 2024 employed TTT-based strategies involving grid perturbations, iterative self-labeling, and contrastive learning to enhance model robustness. These findings reinforce the idea that training solely on static demonstrations is insufficient for high-level reasoning. Models must also be equipped with mechanisms for online adaptation.

In summary, test-time task augmentation represents a powerful tool for improving adaptivity in abstract reasoning tasks. It enables models to dynamically explore and refine hypotheses in response to the structure of the current task, rather than relying only on pretraining or fixed heuristics. Future work should investigate richer augmentation schemes, adaptive loss functions, and hybrid optimization strategies to further leverage the potential of TTT in advancing human-like reasoning systems.
\section{Conclusion}

The Abstraction and Reasoning Corpus (ARC) was designed to probe the foundations of general intelligence by evaluating how well AI systems can solve novel problems with minimal supervision. Unlike traditional benchmarks focused on pattern recognition or data-driven learning, ARC demands conceptual abstraction, structural understanding, and flexible reasoning. These qualities are hallmarks of what has been referred to as \textit{System~2 reasoning}, which goes beyond statistical learning to encompass deliberative, adaptive, and goal-directed cognition.

In this work, we systematically examined the limitations of three dominant paradigms—program synthesis, large language models (LLMs), and transformer-based architectures—when applied to ARC tasks. While each has demonstrated partial success, they remain fundamentally constrained in their ability to generalize to truly novel tasks and adapt dynamically in test-time scenarios. This gap is especially pronounced in adaptation: current systems rarely revise their hypotheses in light of new evidence or shift strategies mid-task. As a result, they fall short of the robust, flexible intelligence required for AGI-level reasoning.

To address these challenges, we proposed three interconnected research directions aimed at promoting both generality and adaptivity in AI models:

First, \textbf{symbolic task abstraction} emphasizes the need for AI systems to extract structured representations that encode high-level semantics from low-level inputs. These representations support compositional reasoning, cross-task generalization, and interpretability, which are essential ingredients for flexible problem solving.

Second, we proposed an \textbf{interactive, feedback-driven reasoning loop}, inspired by the human ability to refine strategies through trial and error. Models equipped with such a loop can adapt their internal hypotheses based on feedback, enabling context-sensitive reasoning and iterative improvement during inference.

Third, we advocated for \textbf{test-time task augmentation}, which leverages task variants or perturbed versions of input at inference time to enhance robustness. This strategy helps models move beyond brittle, distribution-specific performance and toward more adaptable, general-purpose intelligence.

These three components form a blueprint for enabling System~2-style cognition in AI systems. Together, they offer a promising path forward for building models that can generalize structurally, adapt procedurally, and reason strategically, without relying on vast amounts of pretraining data or brittle heuristics.

\textbf{Future work} must extend beyond evaluating models on final answers alone and instead investigate their intermediate reasoning processes, adaptation trajectories, and the structural representations they build internally. Datasets such as ARC, especially when supplemented with human trajectories or richer task variations, provide fertile ground for such inquiry. Furthermore, integrating reinforcement learning, curriculum learning, and neuro-symbolic architectures with these directions may lead to systems that not only imitate human behavior but approach the deeper reasoning faculties that define it.

We contend that System~2 reasoning is not merely a desirable feature but a \textit{necessary foundation} for any model aspiring to AGI. Without structured abstraction, active feedback integration, and robust adaptation, AI will remain confined to narrow domains, regardless of scale. ARC is not just a benchmark but a proving ground for these capabilities. Advancing ARC performance through these principles is not a side path but a direct route to general intelligence. It is time for the field to stop optimizing for superficial benchmarks and begin designing models that think, adapt, and reason. Ultimately, \textbf{we argue that the path to robust, aligned, and general AI does not lie in scaling data or parameters, but in cultivating the mechanisms of System~2 reasoning—abstraction, adaptation, and interactive learning.}

\bibliography{aaai2026}

\end{document}